\documentclass[twoside,bezier,reqno]{amsart}

\usepackage{epsfig}
\usepackage{amsmath,amsthm,amsfonts,amssymb,amscd,euscript}
\usepackage{amsmath,amssymb,amsfonts,amsthm,graphicx}
\usepackage{algorithm}
\usepackage{algpseudocode}
\usepackage{cleveref}
\usepackage{multirow}

\input{epsf}
\newcommand{\filebegin}{\begin{document}}
\newcommand{\fileend}{\end{document}}
\makeatother
\def\thefootnote{}
\newcommand{\lo}{\longrightarrow}
\newcommand{\NMM}{\hspace*{2mm}}

\renewcommand{\baselinestretch}{1.1}

\input{epsf}
\renewcommand{\baselinestretch}{1.1}
\def\n{\noindent}%

\numberwithin{equation}{section}
\def\mapdown#1{\Big\downarrow\rlap
{$\vcenter{\hbox{$\scriptstyle#1$}}$}}
\newtheorem{theorem}{Theorem}[section]
\newtheorem{lemma}[theorem]{Lemma}
\newtheorem{proposition}[theorem]{Proposition}
\newtheorem{corollary}[theorem]{Corollary}

\theoremstyle{definition}
\newtheorem{definition}[theorem]{Definition}
\newtheorem{example}[theorem]{\sc Example}
\newtheorem{xca}[theorem]{Exercise}
\theoremstyle{remark}
\newtheorem{remark}[theorem]{Remark}
\newcommand{\set}[1]{\left\{ #1 \right\}}
\newcommand{\power}[1]{\mathcal{P}(#1)}
\newcommand{\npower}[1]{\mathcal{P}^{\star}(#1)}
\newcommand{\B}{\left\{ 0 , 1 \right\}}
\newcommand{\inflec}[1]{\mathcal{IS}(#1)}
\newcommand{\infp}[2]{\mathcal{IP}_{#1}(#2)}
\newcommand{\sorder}[2]{#1 \preccurlyeq_S #2}
\newcommand{\low}[2]{\underline{#1}(#2)}
\newcommand{\up}[2]{\overline{#1}(#2)}
\newcommand{\ic}[2]{\text{IC}_{#1}(#2)}
\newcommand{\nc}[2]{\text{NC}_{#1}(#2)}
\newcommand{\un}[2]{\text{UN}_{#1}(#2)}
\newcommand{\ps}[2]{\text{POS}_{#1}(#2)}
\newcommand{\nega}[2]{\text{NEG}_{#1}(#2)}
\newcommand{\br}[2]{\text{BR}_{#1}(#2)}
\newcommand{\ind}[2]{\text{IND}_{#1}(#2)}
\newcommand{\bo}[1]{\mathcal{O}(#1)}
\newcommand{\np}{\mathcal{NP}}
\newcommand{\p}{\mathcal{P}}
\newcommand{\pair}[2]{\langle #1, #2 \rangle}
\newcommand{\upg}[1]{\up{G}{#1}}
\newcommand{\lowg}[1]{\low{G}{#1}}
\newcommand{\upgo}[1]{\up{G_{1\oplus2}}{#1}}
\newcommand{\lowgo}[1]{\low{G_{1\oplus2}}{#1}}
\newcommand{\upgp}[1]{\up{G_{1\otimes2}}{#1}}
\newcommand{\lowgp}[1]{\low{G_{1\otimes2}}{#1}}
\newcommand{\gequiv}[2]{#1 \approx_G #2}
\newcommand{\sg}[3]{\text{sg}_{#1}^{#2}\left(#3\right)}
\DeclareMathOperator{\bel}{Bel}
\DeclareMathOperator{\pl}{Pl}
\newcommand{\lowqual}[2]{\underline{Q}_{#1}(#2)}
\newcommand{\upqual}[2]{\overline{Q}_{#1}(#2)}
\renewcommand{\paragraph}[1]{\smallskip\noindent\textsc{#1.}}
\usepackage[symbol]{footmisc}

\renewcommand{\thefootnote}{\fnsymbol{footnote}}

\begin{document}
%%%%%%%%%%%%%%%%%%%%%%%%%%%%%%%%%%%%%%%

\setcounter{page}{1} \noindent
Iranian Journal of Mathematical Sciences and Informatics \\
Vol. x, No. x (201x), pp xx-xx

%%%%%%%%%%%%%%%%%%%%%%%%%%%%%%%%%%%%%%%
\vspace*{2cm}
\begin{center}
{\bf\large A Note on Belief Structures and S-approximation Spaces}
 \\[0.5cm]
{Ali~Shakiba$^{a}$\footnote{Corresponding Author}, Amir~Kafshdar~Goharshady$^{b}$\footnote{Recipient of a DOC fellowship of the Austrian Academy of Sciences}, Mohammad~Reza~Hooshmandasl$^{c}$, Mohsen~Alambardar~Meybodi$^{d}$  \\[2mm]
$^a$Department of Computer Science, Vali-e-Asr University of Rafsanjan, Rafsanjan, Iran\\
$^b$IST Austria (Institute of Science and Technology Austria), Klosterneuburg, Austria\\
$^{c}$Department of Computer Science, University of Mohaghegh Ardabili, Ardabil, Iran\\[2mm]
$^{d}$Department of Applied Mathematics and Computer Science, University of Isfahan, Isfahan, Iran\\[2mm]
{\tt E-mail: ali.shakiba@vru.ac.ir,a.shakiba.iran@gmail.com}\\
{\tt E-mail: amir.goharshady@ist.ac.at}\\
{\tt E-mail: hooshmandasl@uma.ac.ir,hooshmandasl@yazd.ac.ir}\\
{\tt E-mail: m.alambardar@sci.ui.ac.ir}
} \\[2mm]
\end{center}%
\vspace*{0.5cm}
\begin{quotation}
\noindent
{\footnotesize
{\sc Abstract.}
We study relations between evidence theory and S-approximation spaces. Both theories have their roots in the analysis of Dempster's multivalued mappings and lower and upper probabilities, and have close relations to rough sets. We show that an S-approximation space, satisfying a monotonicity condition, can induce a natural belief structure which is a fundamental block in evidence theory. We also demonstrate that one can induce a natural belief structure on one set, given a belief structure on another set, if the two sets are related by a partial monotone S-approximation space. 
}
\end{quotation}

\n \textbf{2000 Mathematics subject classification: } 03B42, 38T37.%\\

\markboth 
{A.~Shakiba, A.K.~Goharshady M.R.~Hooshmandasl, M.~Alambardar~Meybodi}
 {A Note on Belief Structures and S-approximation Spaces}

%%%%%%%%%%%%%%%%%%%%%%%%%%%%%%%%%%%%%%%%%%%%%%%%%%%%%%%%%%%%%%%%%%%%%%%%%%%%%%%%%%%%%%%%%%%%%%%%%%%%%%%%%%%%%%%%%%

%%%%%%%%%%%%%%%%%%%%%%%%%%%%%%%%%%%%%%%%%%%%%%%%%%%%%%%%%%%%%%%%%%%%%%%%%%
\newpage
\section{Introduction}
	\paragraph{Dempster-Shafer Theory of Evidence} The Dempster-Shafer theory of evidence is a well-known method in dealing with uncertainty in problems. It originated in 1967 with the introduction of lower and upper probabilities by Dempster~\cite{Dempster1967_Upperandlower}. A \emph{belief structure} is a fundamental concept in this theory which assigns two numeric values to each subset of a given set. These values are known as the \emph{belief} and \emph{plausibility} measures. See~\cite{Shafer1976} for a detailed treatment. 
	
	\paragraph{S-Approximation} S-approximation spaces are a new way of handling uncertainty, which also originated from Dempster's concepts of lower and upper probabilities~\cite{hindawi}. The motivation for this new approach is that it can be seen as a unifying view to rough sets and their extensions, such as~\cite{Davvaz2008_shortnotealgebraic,Pawlak1982_Roughsets,Pei2004_Roughsetmodels,Skowron1996_Toleranceapproximationspaces,Yao1998_Constructiveandalgebraic,Yao1998_Generalizedroughset,Ziarko1993_Variableprecisionrough}, since they are all expressible in terms of S-approximation spaces~\cite{survey,hindawi}. Hence, any results obtained over S-approximations can be naturally applied to rough sets and many of their extensions, too\footnote{This includes all the results reported in the current paper.}. However, S-approximations are capable of representing more than (extensions of) rough sets and model a very broad range of possible approximations (See  \cite{hindawi,survey} for more examples).
	
	\paragraph{Previous Works on S-Approximations} The concept of S-approximation has been studied by several approaches and its relation to various theories have been examined. For example, S-approximations are studied in the context of Yao's three-way decisions theory~\cite{Yao_2012_Outline,fi} and extended its results. Moreover, they have also been studied in the contexts of neighborhood systems~\cite{Yao1998_Relationalinterpretationsneighborhood,kais}, intuitionistic fuzzy set theory~\cite{jifs} and with relations to topology~\cite{topology}. 
	
	\paragraph{Motivation} Given the common background and overlap of goals, connections between evidence theory and other theories of approximation have been studied for a long time, e.g.~its connections to the theory of rough sets are considered in~\cite{Lin2015_informationfusionapproach,Xu2015_Knowledgereductionsin,Yao1998_Interpretationsbelieffunctions}. The close links between S-approximation spaces and rough sets suggest that a study of relations between evidence theory and S-approximation spaces can yield to more general variants of these results. In this work, we obtain such results about the connections between evidence theory and S-approximation spaces and propose paths for future research.
	
	\paragraph{Organization} The paper is organized as follows: In Section~\ref{sec.premilinaries}, we first review some basic facts from evidence theory, S-approximation spaces, and their corresponding three-way decisions. Then, we study the connection between S-approximation spaces and evidence theory in Section~\ref{sec.s.approximations.belief}. Finally, the paper concludes in Section~\ref{sec.conclusion} by suggesting interesting directions for future research.
	
	\section{Preliminaries}\label{sec.premilinaries}
	
	\subsection{Dempster-Shafer Theory of Evidence}
	In this section, we briefly discuss some background on the Dempster-Shafer theory of evidence. We follow the standard presentation in~\cite{Shafer1976}.
	
	\paragraph{Basic Probability Assignments} A \emph{basic probability assignment}, or \emph{bpa} for short, is a fundamental concept in evidence theory. Let $W$ be a finite non-empty set. Then, a bpa over $W$ is a mapping  $m:\power{W} \rightarrow [0,1]$ satisfying the following conditions: (a)~$m(\emptyset) = 0$, and (b)~$\sum_{X \subseteq W} m(X) = 1$. 
	
	\paragraph{Belief Structures} A set $X \subseteq W$ is called a \emph{focal element} of $m$ if $m(X) \neq 0$. Let $\mathcal{M}$ be the collection of all focal elements of $m$, then the pair $\left( \mathcal{M}, m \right)$ is called a \emph{belief structure} on $W$.
	
	\paragraph{Belief and Plausibility} Given a belief structure $\left( \mathcal{M}, m \right)$, a \emph{belief function} $\bel:\power{W} \rightarrow [0,1]$ and a \emph{plausibility function} $\pl:\power{W} \rightarrow [0,1]$ can be derived, which are defined as follows for every $X \subseteq W$:
	\begin{equation}
		\bel(X) := \sum_{Y \subseteq X} m(Y),
	\end{equation}
	and
	\begin{equation}
		\pl(X) := \sum_{Y \cap X \neq \emptyset} m(Y),
	\end{equation}
	respectively.
	Note that the $\bel$ and $\pl$ functions are duals, i.e.~$\bel(X) = 1 - \pl(X^c)$. Moreover, $[\bel(X),\pl(X)]$ and $\pl(X) - \bel(X)$ are called the \emph{confidence interval} and the \emph{ignorance level} of $X$, respectively.
	
	\paragraph{Axiomatic Approach} A belief function can equivalently be defined in an axiomatic manner, i.e.~it must satisfy the following axioms:
	\begin{itemize} 
		\item $\bel(\emptyset) = 0$, 
		\item $\bel(W) = 1$, 
		\item $\bel \left( \cup_{i = 1}^{\ell} X_i \right) \geq \sum_{\emptyset \neq I \subseteq \set{1,\ldots,\ell}} (-1)^{|I| + 1} \bel(\cap_{i \in I} X_i)$ for $\set{X_1,\ldots,X_{\ell}} \subseteq \power{W}$ and $\ell > 0$.
	\end{itemize}
	
	\subsection{S-approximation spaces}
	In this section, some basic facts and definitions for S-approximation spaces are presented. We follow the notation of~\cite{fi,hindawi}.
	
	\paragraph{S-Approximation Spaces} An \emph{$S$-approximation space} is formally defined as a quadruple $G = (U, W, T, S)$, where $U$ and $W$ are finite non-empty sets, $T$ is a multi-valued mapping $T: U \rightarrow \power{W}$, called a \emph{knowledge component}, and $S$ is a mapping  $S : \power{W} \times \power{W} \rightarrow \left\{ 0 , 1 \right\}$, called a \emph{decider}.
	
	\paragraph{Lower and Upper Approximations} Given an $S$-approximation space $G = (U, W, T, S)$, the lower and upper approximations of $X \subseteq W$ are defined as
	\begin{equation}
		\underline{G}(X) = \left\{ x \in U \ \vert \  S(T(x), X) = 1 \right\} ,
	\end{equation}
	and
	\begin{equation}
		\overline{G}(X) = \left\{ x \in U \ \vert \  S(T(x), X^c) = 0 \right\} ,
	\end{equation}
	respectively, where $X^c$ denotes the complement of $X$ with respect to $W$. 
	
	\paragraph{Generality and Special Cases} Note that the mapping $S$ can model a large class of measures, of which set inclusion, i.e.~$S_\subseteq(A, B) = \left\{\begin{matrix}
	1 & A \subseteq B\\ 
	0 & \text{otherwsie} 
	\end{matrix}\right.$, is a special case. If we set $S$ to $S_\subseteq$ and consider the sets of form $T(x)$ as blocks, we can model rough sets and some of their generalizations as special cases. For more information and other examples of decider functions consult \cite{hindawi,fi,kais,survey}. Moreover, other definitions and extensions have also been proposed for decider mappings, e.g. refer to~\cite{jifs} to see an instance suitable for intuitionistic fuzzy sets. However, in this paper we stick to the standard and general definition of S-approximation spaces as defined above. 
	
	\paragraph{Trichotomy Regions} For any set $X \subseteq W$, the three pair-wise disjoint sets of positive, negative and boundary regions are defined as follows:
	\begin{equation}
		\label{eq.three.regions.ordinary}
		\begin{array}{rlr}
			\ps{G}{X} := & \set{ x \in U\ \vert \ S(T(x), X) = 1 \wedge S(T(x), X^c) = 0} & \text{(Positive Region)}\\
			= & \lowg{X} \cap \upg{X}, &  \\
			\nega{G}{X} := & \set{ x \in U\ \vert \ S(T(x), X) = 0 \wedge S(T(x), X^c) = 1} & \text{(Negative Region)} \\
			= & U \setminus \left( \lowg{X} \cup \upg{X} \right), &  \\
			\br{G}{X} := & \set{ x \in U\ \vert \ S(T(x), X) = S(T(x), X^c)} & \text{(Boundary Region)} \\
			= & \lowg{X} \Delta \upg{X}, & 
		\end{array}
	\end{equation}
	where $A \Delta B = (A \setminus B) \cup (B \setminus A)$ for $A,B \subseteq U$.
	
	It is noteworthy that the intuition behind Equation \ref{eq.three.regions.ordinary} is very similar to that of~\cite{Yao_2012_Outline} (Equation 1). Refer to~\cite{kais} for more discussion on this point. It is also the case that $\ps{G}{X} = \nega{G}{X^c}$ and $\br{G}{X} = \br{G}{X^c}$ for any $X \subseteq W$ \cite{kais}. We will routinely use these facts throughout the paper.
	
	\paragraph{Partial Monotonicity} A decider mapping $S: \power{W} \times \power{W} \to \B$ is called \emph{partial monotone} if $X \subseteq Y \subseteq W$ and $S(A,X) = 1$ imply that $S(A,Y) = 1$ for any $A \subseteq W$. An S-approximation space $G=(U,W,T,S)$ with a partial monotone decider mapping $S$ is called a partial monotone S-approximation space. The lower and upper approximation operators and the three decision regions of such S-approximation spaces satisfy several important properties which are listed in the following proposition:
		
		\begin{proposition}[\cite{fi,survey}]
			\label{prop.main.props}
			%			\cite{fi} 
			Let $G=(U, W, T, S)$ be a partial monotone S-approximation space. For all $X, Y \subseteq W$, we have:
			\begin{enumerate}
				\item $ X \subseteq Y $ implies $\overline{G}(X) \subseteq \overline{G}(Y) $,
				\item $ X \subseteq Y $ implies $\underline{G}(X) \subseteq \underline{G}(Y)$,
				\item $ \overline{G}(X \cup Y) \supseteq \overline{G}(X) \cup \overline{G}(Y)$,
				\item $ \overline{G}(X \cap Y) \subseteq \overline{G}(X) \cap \overline{G}(Y) $,
				\item $ \underline{G}(X \cup Y) \supseteq \underline{G}(X) \cup \underline{G}(Y) $,
				\item $ \underline{G}(X \cap Y) \subseteq \underline{G}(X) \cap \underline{G}(Y) $,
				\item $ \overline{G}(X) = (\underline{G}(X^c))^c $,
				\item $ \underline{G}(X) = (\overline{G}(X^c))^c $,
				\item $X \subseteq Y$ implies $\ps{G}{X} \subseteq \ps{G}{Y}$,
				\item $X \subseteq Y$ implies $\nega{G}{Y} \subseteq \nega{G}{X}$,
				\item $\ps{G}{X \cup Y} \supseteq \ps{G}{X} \cup \ps{G}{Y}$,
				\item $ \nega{G}{X \cup Y} \subseteq \nega{G}{X} \cup \nega{G}{Y} $,
				\item $\ps{G}{X \cap Y} \subseteq \ps{G}{X} \cap \ps{G}{Y}$,
				\item $\nega{G}{X \cap Y} \supseteq \nega{G}{X} \cap \nega{G}{Y}$,
				\item $\ps{G}{X} \cap \nega{G}{Y} \subseteq \ps{G}{X} \cap \nega{G}{X \cap Y}$.
			\end{enumerate}
		\end{proposition}
		
		\paragraph{Inflection Sets} Partial monotone S-approximation spaces can be represented by an equivalent form, which is called an \emph{inflection set}. A pair $(x,X) \in U \times \power{W}$ is called an \emph{inflection point} with respect to $G$ whenever $S(T(x), X) = 1$ and for all $Y \subsetneq X$, we have $S(T(x), Y)=0$ \cite{fi}. The inflection set of a partial monotone $G$, which is denoted by $\inflec{G}$, is defined as the set of all of its inflection points. Moreover, for $x \in U$ we use $\infp{G}{x}$ to represent the collection of $X \subseteq W$ where $(x,X) \in \inflec{G}$, so that $\inflec{G} = \cup_{x \in U} \set{(x,X) \vert X \in \infp{G}{x}}$.
		
		\paragraph{Trivial Elements} An element $x \in U$ is called \emph{trivial} if we have either $\infp{G}{x} = \emptyset$ or $\infp{G}{x} = \set{\emptyset}$. In the former case, we have $S(T(x), X) = 0$ for all $X \subseteq W$, so $x$ appears in none of the lower approximations $\underline{G}(X)$ and in every upper approximation $\overline{G}(X^c)$. So, the element $x$ is not providing any useful information, i.e.~it cannot be used to distinguish any pair of subsets of $W$. Similarly, in the latter case, $S(T(x), \emptyset) = 1$, which, due to partial monotonicity, implies $S(T(x), X) = 1$ for all $X \subseteq W$. Hence, for all $X \subseteq W$, we have $x \in \underline{G}(X)$ and $x \not\in \overline{G}(X^c)$. So $x$ does not provide any useful information in this case, either.
		
		\paragraph{Reducibility} As argued above, if $x$ is a trivial element, one can remove $x$ and get a smaller system from which one can get just as much information as the initial system. A partial monotone S-approximation space is called \emph{reducible} if it contains a trivial element, otherwise we call it \emph{irreducible}. 
		
		%		A \emph{complement compatible} S-approximation space $G=(U,W,T,S)$ is the one which for all $x \in U$ and $X \subseteq W$, either $S(T(x), X) = 0$ or $S(T(x), X^c) = 0$. For these S-approximation spaces, we have $\underline{G}(X) \subseteq \overline{G}(X)$ \cite{fi}.
		
		\section{S-approximation spaces and belief structures}\label{sec.s.approximations.belief}
		In this section, we study the relationship between S-approximation spaces and belief structures.
		
		The qualities of lower and upper approximations with respect to an S-approximation space are defined as follows:
		\begin{definition}
			\label{def.quality.lower.upper}
			Let $G=(U,W,T,S)$ be an $S$-approximation space. The qualities of lower and upper approximations of a set $X \subseteq W$ with respect to $G$ are defined as:
			\begin{equation}
				\label{eq.quality.lower}
				\lowqual{G}{X} = \frac{\left| \ps{G}{X} \right|}{\left| U \right|},
			\end{equation}
			and
			\begin{equation}
				\label{eq.quality.upper}
				\upqual{G}{X} = \frac{\left| \ps{G}{X} \right| + \left| \br{G}{X} \right|}{\left| U \right|}.
			\end{equation}
		\end{definition}
		The qualities defined in Equations \ref{eq.quality.lower} and \ref{eq.quality.upper} are dual. This is stated more formally in the following proposition:
		\begin{proposition}
			Let $G=(U,W,T,S)$ be an $S$-approximation space. Then, for all $X \subseteq W$ we have $\lowqual{G}{X} = 1 - \upqual{G}{X^c}$.
		\end{proposition}
		\begin{proof}
			The proof is as follows and uses the fact that $\ps{G}{X} = \nega{G}{X^c}$:
			\begin{equation}
				\begin{aligned}
					\lowqual{G}{X} = & \frac{\left| \ps{G}{X} \right|}{\left| U \right|} = \frac{\left| \nega{G}{X^c} \right|}{\left| U \right|} \\
					= & \frac{\left| U \setminus \left( \ps{G}{X^c} \cup \br{G}{X^c} \right) \right|}{\left| U \right|} \\
					= & 1 - \frac{\left| \ps{G}{X^c} \right| + \left| \br{G}{X^c} \right|}{\left| U \right|} \\
					= & 1 - \upqual{G}{X^c}.
				\end{aligned}
			\end{equation}
			%			The proof is clear from the definition and the fact that $\ps{G}{X} = \nega{G}{X^c}$.
			%\qed 
		\end{proof}
		Next, we consider the properties of these quality values for a partial monotone S-approximation space.
		\begin{proposition}
			\label{lem.bel.pm.empty}
			Let $G=(U,W,T,S)$ be a partial monotone $S$-approximation space. Then, $\lowqual{G}{\emptyset} = 0$.
		\end{proposition}
		\begin{proof}
			It suffices to show that $\ps{G}{\emptyset} = \emptyset$. The proof is by contradiction. Suppose there exists some $x \in U$ such that $x \in \ps{G}{\emptyset}$. So, it is the case that $S(T(x), \emptyset) = 1$ and $S(T(x), W) = 0$. This is a contradiction with partial monotonicity of $G$, since $\emptyset \subseteq W$ and we need to have $S(T(x), W) = 1$. Therefore the desired result is obtained.
			%			The proof is clear.
			%\qed
		\end{proof}
		\begin{proposition}
			\label{lem.bel.pm.w}
			Let $G=(U,W,T,S)$ be an irreducible partial monotone $S$-approximation space. Then, $\lowqual{G}{W} = 1$.
		\end{proposition}
		\begin{proof}
			Note that $G$ is irreducible, hence for every $x \in U,$ there exists $X \subseteq W,$ such that $S(T(x), X) = 1$\footnote{Otherwise $x$ is trivial and $G$ is reducible, which is a contradiction.}. Therefore, by partial monotonicity, we have $S(T(x), W) = 1$ for all $x \in U$. Moreover, $S(T(x), \emptyset) = 0$ for all $x \in U$. Hence, $x \in \ps{G}{W}$ for all $x \in U$ and $\ps{G}{W} = U$. So, $\lowqual{G}{W} = \tfrac{|U|}{|U|} = 1$.
			%\qed
		\end{proof}
		\begin{proposition}
			\label{lem.bel.pm.tri.ineq}
			Let $G=(U,W,T,S)$ be a partial monotone $S$-approximation space. Then, for all $\ell \in \mathbb{N}$ we have
			\begin{equation}
				\lowqual{G}{\cup_{i=1}^{\ell} X_i} \geq \sum_{\emptyset \neq I \subseteq \set{1,\ldots,\ell}}(-1)^{|I|+1} \lowqual{G}{\cap_{i \in I} X_i},
			\end{equation}
			where $X_i \subseteq W$.
		\end{proposition}
		\begin{proof}
			By the definition, we have
			\begin{equation}
				\lowqual{G}{\cup_{i=1}^{\ell} X_i} = \frac{\left| \ps{G}{\cup_{i=1}^{\ell} X_i} \right|}{|U|}.
			\end{equation}
			By partial monotonicity of $G$, we have
			\begin{equation}
				\frac{\left| \ps{G}{\cup_{i=1}^{\ell} X_i} \right|}{|U|} \geq \frac{\left| \cup_{i=1}^{\ell} \ps{G}{X_i} \right|}{|U|},
			\end{equation}
			since $\ps{G}{\cup_{i=1}^{\ell} X_i} \supseteq \cup_{i=1}^{\ell} \ps{G}{X_i}$. Now the desired result can be obtained by appling the inclusion-exclusion principle.
			%\qed
		\end{proof}
		Propositions \ref{lem.bel.pm.empty} to \ref{lem.bel.pm.tri.ineq} result in the following:
		\begin{proposition}
			\label{cor.pm.s.app.to.belief}
			Let $G=(U,W,T,S)$ be an irreducible partial monotone $S$-approximation space. The quality of lower approximation, as defined in Definition~\ref{def.quality.lower.upper}, is a belief function.
		\end{proposition}
		Similarly, for an irreducible partial monotone S-approximation space, the quality of upper approximation is a plausibility function. This is treated more formally in the following proposition:
		\begin{proposition}
			\label{prop.pm.is.pl.upper.qual}
			Let $G=(U,W,T,S)$ be an irreducible partial monotone $S$-approximation space. Then the quality of upper approximation, as defined in Definition~\ref{def.quality.lower.upper}, is a plausibility function.
		\end{proposition}
		\begin{proof}
			By the duality of belief and plausibility functions, we have $\pl_G(X) = 1 - \bel_G(X^c)$ and this is all we have to show. By the definition, we have 
			\begin{equation}
				\begin{aligned}
					\lowqual{G}{X^c} = & \frac{\left| \ps{G}{X^c} \right|}{\left| U \right|} \\
					= & \frac{\left| \nega{G}{X} \right|}{\left| U \right|}  \\
					= & \frac{\left| U \setminus \left( \ps{G}{X} \cup \br{G}{X} \right) \right|}{|U|} \\
					= & 1 - \frac{\left| \ps{G}{X} \right| + \left| \br{G}{X} \right|}{|U|} \\
					= & 1 - \upqual{G}{X}.
				\end{aligned}
			\end{equation}
			By applying Proposition \ref{cor.pm.s.app.to.belief}, the desired result is obtained.
		\end{proof}
		By Propositions \ref{cor.pm.s.app.to.belief} and \ref{prop.pm.is.pl.upper.qual}, it can be said that every irreducible partial monotone S-approximation space induces a belief structure on $W$.
		\begin{theorem}
			Let $G=(U,W,T,S)$ be an irreducible partial monotone $S$-approximation space. Then, $G$ induces a belief structure $\left( \mathcal{M},m \right)$ on $W$ where
			\begin{equation}
				m(X) = \sum_{Y \subseteq X} \left( -1 \right)^{\left| X \setminus Y \right|} \lowqual{G}{Y},
			\end{equation}
			and
			\begin{equation}
				\mathcal{M} = \set{X \subseteq W \ \vert \ m(X) \neq 0},
			\end{equation}
			for $X \subseteq W$.
		\end{theorem}
		\begin{proof}
			The bpa can be defined from a belief function, which is the quality of lower approximation (by Proposition \ref{cor.pm.s.app.to.belief}), by the following relation 
			\begin{equation}
				n(A) = \sum_{B \subseteq A} (-1)^{\left| A \setminus B\right|}  \bel(B),
			\end{equation}
			where $n$ is a bpa \cite{Yao1998_Interpretationsbelieffunctions}. This concludes the proof.
		\end{proof}
		
		Next, we show that belief structures can induce S-approximation spaces. 
		\begin{theorem}
			\label{thm.s.approx.induced.belief}
			Suppose that $\left(\mathcal{M},m\right)$ is a given belief structure over a finite non-empty set $W$ such that for all focal elements $X \in \mathcal{M}$, there exist  $a,b \in \mathbb{Z}^{+}$ such that $m(X) = \frac{a}{b}$. Then, there exists an S-approximation space $G=(U,W,T,S)$ such that the quality of lower and upper approximations with respect to $G$ are the corresponding belief and plausibility functions.
		\end{theorem}
		\begin{proof}
			The proof is by construction. Without loss of generality, we assume that there exists a constant $d \in \mathbb{Z}^{+}$ such that for all focal elements $X \in \mathcal{M}$, we have $m(X) = \frac{c}{d}$ for some $c \in \mathbb{Z}^{+}$. This is easy to obtain by computing the least common multiple.
			
			Now define the set $U$ as $U=\set{1,\ldots,d}$.
			 For each $X \in \mathcal{M}$ with $m(X) = \frac{l_X}{d}$, we choose a subset $A_X$ of size $l_X$ of $U$. We assume that the $A_X$'s are pairwise disjoint. We can always find such disjoint $A_X$'s, since $\sum_{X \in \mathcal{M}} m(X) = 1$ and hence $\sum_{X \in \mathcal{M}} l_X = d.$ Now for each $i \in A_X$, we let $T(i) = X$. Finally, we let the decider mapping $S$ be the ordinary set inclusion operator $S_\subseteq$.
			
			Next, it is easy to see that $G$ satisfies the conditions of Propositions~\ref{cor.pm.s.app.to.belief} and~\ref{prop.pm.is.pl.upper.qual}. Therefore, the qualities of lower and upper approximations with respect to $G$ are belief and plausibility functions, respectively.
			
			Finally, we show that for all $X \subseteq W$, the belief and plausibility values of $X$ with respect to $\left( \mathcal{M}, m \right)$ are equal to the corresponding values with respect to $G$. Since the belief and plausibility functions are dual, it suffices to show the result for belief. This can be done as follows:
			
			\begin{equation}
				\begin{aligned}
					\lowqual{G}{X} = & \frac{\left| \ps{G}{X} \right|}{\left| U \right|} \\ 
					= & \frac{\left| \set{x \in U \ | \ T(x) \subseteq X} \right|}{\left| U \right|} \\
					= & \sum_{Y \subseteq X} m(Y) = \bel(X).
				\end{aligned}
			\end{equation}
			This concludes the proof.
		\end{proof}
		
		Now suppose that we are given a belief structure $(\mathcal{M}, m)$ over $U$ and an irreducible partial monotone S-approximation space $G=(U,W,T,S)$. Then we can induce a belief structure $(\mathcal{M}', m')$ on $W$ by declaring $\mathcal{M}'$ as
		\begin{equation}
			\label{eq.m.w.focal}
			\mathcal{M}' = \set{Z \subseteq W \ \vert \ \exists x \in U, (x,Z) \in \inflec{G}}\!,
		\end{equation}
		and the bpa $m'$ as
		\begin{equation}
			\label{eq.m.w.bpa}
			m'(Y) = \left\{
			\begin{array}{lll}
				\sum_{X \in \mathcal{M}}  \frac{m(X)}{\left| X \right|} \times \left( \sum_{x \in X,~ Y \in \infp{G}{x}} \frac{1}{\left| \infp{G}{x} \right|} \right)  & \text{if} & Y \in \mathcal{M}', \\
				0 & \multicolumn{2}{l}{\text{otherwise}.}
			\end{array}
			\right.
		\end{equation}
		The intuition behind Equation \ref{eq.m.w.bpa} is that the bpa value of every $X \in \mathcal{M}$ is divided between each $x \in X$ equally likely, which are called their shares. Then, the bpa $m'$ of $Y \in \mathcal{M}'$ receives the shares of those $x \in X$ for which $Y \in \infp{G}{X}$. 
		\begin{theorem}
			Given a belief structure $(\mathcal{M},m)$ on a finite non-empty set $U$ and an irreducible partial monotone S-approximation space $G=(U,W,T,S)$, $(\mathcal{M}', m')$ as defined in Equations \ref{eq.m.w.focal} and \ref{eq.m.w.bpa} is a valid belief structure on $U$.
		\end{theorem}
		\begin{proof}
			The bpa $m'$ needs to satisfy two conditions, i.e. (1) $m'(\emptyset) = 0$ and (2) $\sum_{Y \subseteq W} m'(Y) = 1$. By the hypothesis that $\emptyset \notin \infp{G}{x}$ for all $x \in U$, we have $\emptyset \notin \mathcal{M}'$ and therefore, its bpa value $m'(\emptyset)$ is zero. The second property can be proven as follows (note that for all $x \in X \in \mathcal{M}$, we have $\infp{G}{x} \subseteq \mathcal{M}'$):
			
			\begin{align}
				\begin{aligned}
					\sum_{Y \subseteq W} m'(Y) = & \sum_{Y \in \mathcal{M}'} m'(Y) \\
					= & \sum_{Y \in \mathcal{M}'} \sum_{X \in \mathcal{M}} \frac{m(X)}{\vert X \vert} \times \left( \sum_{x \in X~\vert~ Y \in \infp{G}{x}} \frac{1}{\vert \infp{G}{x} \vert} \right)\\
					= & \sum_{x \in X \in \mathcal{M}, Y \in \infp{G}{x} \subseteq \mathcal{M}'} \frac{m(X)}{\vert X \vert} \times \frac{1}{\vert \infp{G}{x} \vert}\\
					= & \sum_{X \in \mathcal{M}} \frac{m(X)}{\vert X \vert } \times \left( \sum_{x \in X,~ Y \in \infp{G}{x}} \frac{1}{\vert \infp{G}{x} \vert} \right)\\
					= & \sum_{X \in \mathcal{M}} \frac{m(X)}{\vert X \vert} \times \left(\sum_{x \in X} 1 \right)\\
					= & \sum_{X \in \mathcal{M}} m(X) = 1.
				\end{aligned}
			\end{align}
		\end{proof}

		\section{Conclusion and future research directions}\label{sec.conclusion}
		In this paper, we studied some connections between the Dempster-Shafer's theory of evidence and the concept of S-approximation spaces. First, we  defined two numeric measures called the qualities of lower and upper approximations for $S$-approximation spaces. Then, we showed that they can be used to derive a belief structure from an irreducible partial monotone S-approximation space in a natural way. Finally, we showed that given a belief structure on a set $U$ and an irreducible partial monotone S-approximation space $G=(U,W,T,S)$, a valid natural belief structure can be induced on $W$. 
		
		The results obtained in this paper are the first ones settling a relation between the two theories and are extensible by trying to answer the following proposed problems:
		\begin{enumerate}
			\item Can belief structures be generalized to two universal sets with respect to an arbitrary S-approximation space in a natural or meaningful way?
			\item Can the results of this paper be extended to neighborhood systems, especially the ones in \cite{kais}? For example, by fusing knowledge mappings of multiple S-approximation spaces with a similar approach to \cite{Lin2015_informationfusionapproach}.
			\item Can the qualities of lower and upper approximations be used to reduce the knowledge mappings in the context of \cite{kais}? For example, can one find a minimal set of knowledge mappings of multiple S-approximation spaces for which the amount of information one can obtain from that set does not change compared to the case when she uses all of them?
		\end{enumerate}
	
	\section*{Acknowledgments}
	
	We are very grateful to the anonymous reviewer for detailed comments and suggestions that significantly improved the presentation of this paper. The research was partially supported by a DOC fellowship of the Austrian Academy of Sciences.
		
%\bibliographystyle{plain}
%\bibliography{bibliography}

\end{document}